\documentclass{article}

\PassOptionsToPackage{numbers, compress}{natbib}

\usepackage[preprint]{neurips_2022}

\usepackage[table]{xcolor}
\usepackage{graphicx}
\usepackage[utf8]{inputenc} 
\usepackage[T1]{fontenc}    
\usepackage{hyperref}       
\usepackage{url}            
\usepackage{booktabs}       
\usepackage{amsfonts}       
\usepackage{nicefrac}       
\usepackage{amsmath}
\usepackage{tabulary}
\usepackage{amssymb}
\usepackage{graphicx}
\usepackage{multicol, blindtext}
\usepackage{multirow}
\usepackage{microtype}      
\usepackage{xcolor}         
\newcommand\eg{\emph{e.g.,}} 
\newcommand\ie{\emph{i.e.}}

\newcommand{\Rows}[1]{\multirow{2}{*}{#1}}
\def\Vec#1{{\boldsymbol{#1}}}

\usepackage{bbding}
\usepackage{wrapfig,lipsum,booktabs}
\usepackage{subfig}
\usepackage{pifont}%
\newcommand{\cmark}{\ding{51}\xspace}%
\newcommand{\xmarkg}{\textcolor{lightgray}{\ding{55}}\xspace}%
\newcommand{\gc}[1]{\textcolor{deemph}{#1}}
\definecolor{deemph}{gray}{0.6}
\definecolor{green}{HTML}{39b54a}  
\definecolor{greenn}{HTML}{008000}  
\definecolor{blue}{HTML}{1F51FF}  %

\newcommand{\bl}[1]{\textcolor{blue}{#1}}
\newcommand{\re}[1]{\textcolor{red}{#1}}
\newlength\savewidth\newcommand\shline{\noalign{\global\savewidth\arrayrulewidth
  \global\arrayrulewidth 1pt}\hline\noalign{\global\arrayrulewidth\savewidth}}
\newcommand{\tablestyle}[2]{\setlength{\tabcolsep}{#1}\renewcommand{\arraystretch}{#2}\centering\footnotesize}
\newcolumntype{x}[1]{>{\centering\arraybackslash}p{#1pt}}
\newcolumntype{y}[1]{>{\raggedright\arraybackslash}p{#1pt}}
\newcolumntype{z}[1]{>{\raggedleft\arraybackslash}p{#1pt}}

\definecolor{baselinecolor}{gray}{.9}
\newcommand{\baseline}[1]{\cellcolor{baselinecolor}{#1}}

\title{CropMix: Sampling a Rich Input Distribution via Multi-Scale Cropping}

\author{Junlin Han$^{1,2,3}$, Lars Petersson$^{1}$, Hongdong Li$^{2}$, Ian Reid$^{3}$\\
$^{1}$DATA61-CSIRO, $^{2}$Australian National University,  $^{3}$University of Adelaide \\
junlin.han@data61.csiro.au, lars.petersson@data61.csiro.au,
\\
hongdong.li@anu.edu.au, ian.reid@adelaide.edu.au
}

\begin{document}

\maketitle

\begin{abstract}
We present a simple method, CropMix, for the purpose of producing a rich input distribution from the original dataset distribution.
Unlike single random cropping, which may inadvertently capture only limited information, or irrelevant information, like pure background, unrelated objects, etc, we crop an image multiple times using distinct crop scales, thereby ensuring that multi-scale information is captured. The new input distribution, serving as training data, useful for a number of vision tasks, is then formed by simply mixing multiple cropped views. We first demonstrate that CropMix can be seamlessly applied to virtually any training recipe and neural network architecture performing classification tasks. CropMix is shown to improve the performance of image classifiers on several benchmark tasks across-the-board without sacrificing computational simplicity and efficiency. Moreover, we show that CropMix is of benefit to both contrastive learning and masked image modeling towards more powerful representations, where preferable results are achieved when learned representations are transferred to downstream tasks. Code is available at {\href{https://github.com/JunlinHan/CropMix}{\re{GitHub}}}.

\end{abstract}
\section{Introduction}
Since the introduction of AlexNet~\cite{alexnet} in 2012, deep neural networks have dominated almost all computer vision tasks~\cite{FCN,deeplab,he2017mask,RCNN,vgg,resnet}. Despite the advances, a common phenomenon today is that even a small network can easily overfit the training data, causing a large performance drop and decreased generalization to unseen data~\cite{resnet}. One possible solution is to collect numerous training samples, but both collecting and labeling are prohibitively costly. Even worse, neural networks may still suffer from overfitting when trained using millions of data~\cite{imagenet}. This issue can be mitigated by applying data augmentations to create various transformed views per training sample. Among all data augmentations, cropping is considered the most essential one for classification tasks~\cite{googlenet,resnet}, for the reason that cropping can efficiently sample the network's input distribution from the original dataset distribution. The Random Resized Crop (RRC) introduced by GoogleNet~\cite{googlenet} has been widely adopted and serves as a common practice for ImageNet~\cite{imagenet} classification tasks. 

\begin{figure}[htb]
     \centering
     \includegraphics[width = 13cm]
     {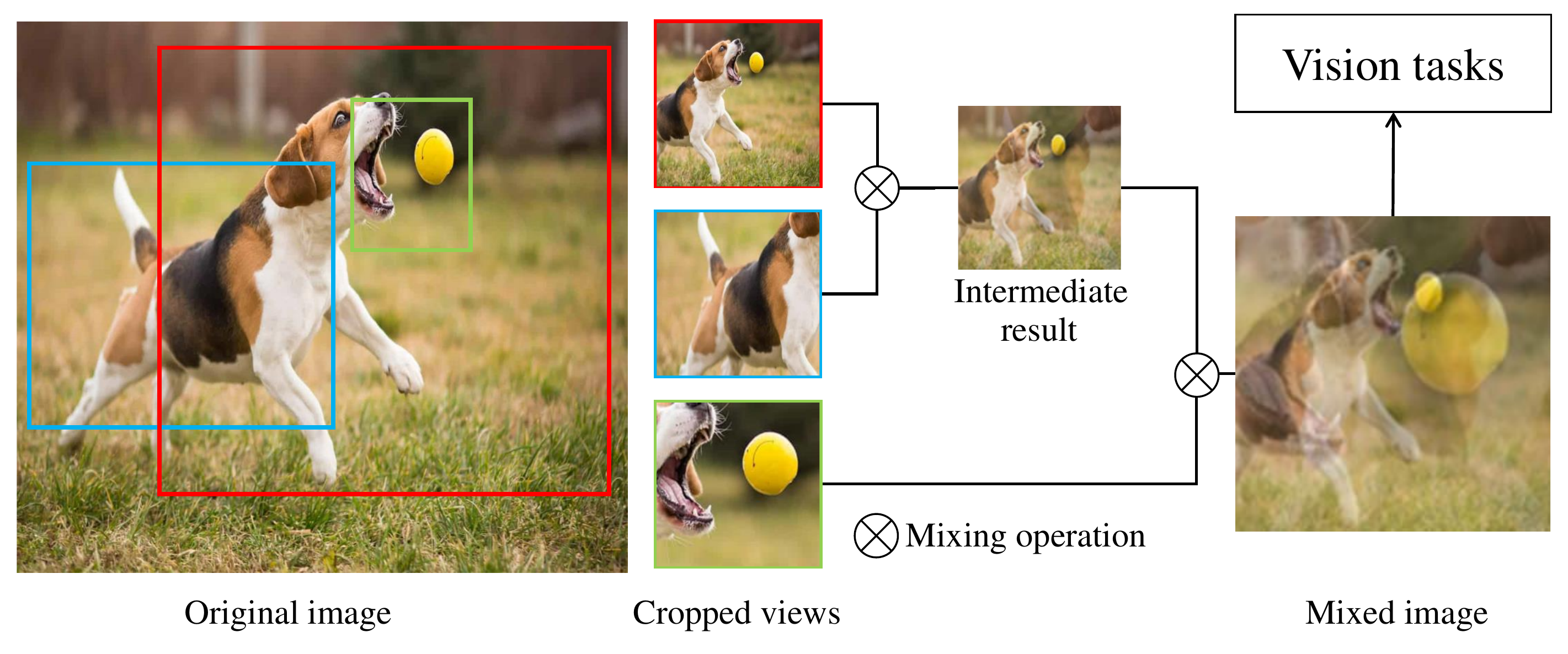}
     \caption{\textbf{A realization of our CropMix}. We utilize multiple cropping operations with distinct crop scales to obtain multiple cropped views. Gradually mixing these cropped views produces a mixed image containing multi-scale information. Applying CropMix allows us to sample a richer input distribution from the dataset distribution while suppressing the risk of labeling errors.
     }
     \label{fig:ppline}
     \vspace{-0.5cm}
\end{figure}

RRC is simple yet highly effective. It crops a random region of the training sample, and then resizes the cropped region to a fixed resolution as the input to neural networks. RRC brings more variability and diversity to the raw sample, due to the dynamic changing of the aspect ratio and the apparent size of objects~\cite{touvron2019fixing}. However, this is not a free lunch. A larger range of crop scale offers more diversity to regularize overfitting, but at the risk of introducing labeling errors~\cite{touvron2022deit}. For instance, when the target object is not within the cropped region, the cropped view might introduce noise (\eg{} irrelevant objects or background) to the label. Also, RRC can only capture single-scale information for each crop. As a result, using a fine-grained scale means sacrificing information of a coarse-grained scale, and vice versa. Considering these issues, some questions arise naturally: Can we increase the diversity and variability without introducing labeling noise? Can we preserve image information from both the local fine-grained scale and the global coarse-grained scale? How can we sample a richer input distribution to represent the original dataset distribution?

In this paper, we propose a simple data pre-processing pipeline, referred to as CropMix, aiming to 1: crop and aggregate multi-scale information, 2: suppress labeling noise/errors, and 3: sample a richer input distribution. The proposed CropMix is a two-step process, including multi-scale cropping followed by mixing. Applying CropMix, where one image is cropped using multiple RRC with distinct crop scales, can obtain multiple cropped views of the same image. In each applied CropMix, RRC with a small crop scale captures fine details, while RRC with a large crop scale covers a greater fraction of the image pixels, such that multi-scale information is captured. In the mixing stage, those cropped views are simply mixed using Mixup~\cite{zhang2017mixup} or CutMix~\cite{yun2019cutmix} to form a new mixed image, preserving the multi-scale information captured from the original data. CropMix always applies RRC with a large crop scale, so the risk of labeling errors can be efficiently suppressed. Moreover, by mixing multiple cropped views, the diversity of the input data is further increased. That is, the input distribution can be richer to better reflect the original dataset distribution. 
With little computational cost~\footnote{Approximately 1.6\% extra training time on ImageNet classification tasks using the ResNet50 model and 8 RTX 2080 Ti GPUs. }, CropMix provides a non-marginal gain in performance among multiple vision tasks.

Our key insight is that \textit{cropping multiple times with distinct crop scales and then mixing can form a richer input distribution representing the original dataset distribution to benefit multiple vision tasks}. We conduct extensive evaluations of CropMix on various network architectures, training recipes, datasets, and challenging tasks. We start our experiments with \textbf{7} network architectures on CIFAR-10 and CIFAR-100~\cite{krizhevsky2009learning} classification tasks with multiple training settings. When the range of crop scales is large, CropMix can greatly outperform RRC in terms of classification accuracy. Next, we conduct experiments on ImageNet-1K~\cite{imagenet} (henceforth referred to as ImageNet), where \textbf{3} training recipes and \textbf{4} network architectures are evaluated. Among \textbf{5} classifier performance metrics, CropMix simultaneously outperforms RRC, sometimes by large margins.

We then evaluate the effectiveness of CropMix on contrastive learning~\cite{hadsell2006dimensionality,chen2020simple,he2020momentum}. Following Asym-Siam~\cite{wang2022asym}, we apply CropMix to the source branch (query), that is, we craft new views of an image by mixing two views created by RRC with distinct scales. Similar to the effect of applying ContrastiveCrop~\cite{peng2022contrastivecrop}, CropMix applied to contrastive learning avoids false negatives (object \textit{vs.} background), and also brings extra variance to the encoding space~\cite{wang2022asym}, ending up with a more powerful representation that can be better utilized by down-stream tasks.

Recently, masked image modeling approaches have raised hopes for a BERT~\cite{devlin2018bert} moment in computer vision~\cite{touvron2022deit}. SimMIM~\cite{xie2021simmim}, BEiT~\cite{bao2021beit}, and MAE~\cite{MaskedAutoencoders2021} all unveil that masked image modeling does not require heavy augmentations, and simple RRC is often sufficient to sample the input distribution for the reconstruction objective. However, we show that sampling a richer input distribution for the reconstruction objective is still beneficial, that is, applying CropMix can help such a self-supervised learning paradigm 
to learn more meaningful knowledge, especially when data is limited.

CropMix is learning-free, easy to employ, and scales well to multiple vision tasks. To understand the effectiveness and intuitions behind it, we comprehensively study and analyze more complex CropMix settings and perform thorough ablation studies.

\section{Related Work}

\textbf{Cropping} is an important data processing step for training neural networks on visual data. The cropping operation can efficiently avoid model overfitting while encouraging that the prediction of the model is invariant to transformations~\cite{touvron2019fixing}. Cropping also helps networks to learn the cognitive ability of recognizing partial patterns~\cite{han2022yoco}. The random cropping with resizing operation changes the apparent size of objects in the images, forcing networks to have a predictable response to scale changes~\cite{touvron2019fixing}. The Random Resized Crop (RRC)~\cite{googlenet}, verified to work efficiently in the ILSVRC competition~\cite{imagenet}, has become a common practice for ImageNet classification tasks. Compared to RRC, CropMix gives more variability to the input images by aggregating multi-scale information, so that a richer input distribution can be sampled from the original dataset distribution. For classification tasks, as views cropped with large crop scales are usually label-preserving, the risk of labeling errors can be efficiently suppressed by applying CropMix. 

\textbf{Mixup}~\cite{zhang2017mixup,inoue2018data} creates mixed input images with soft labels (convex combinations) for training, such that one training sample contains information from multiple images. Multiple variants~\cite{yun2019cutmix,hendrycks2019augmix, verma2019manifold,kim2020puzzle} of Mixup are proposed to create mixed input images with better combinations. Mixup and CutMix~\cite{yun2019cutmix} are widely adopted in training classification networks, especially in training vision transformers~\cite{dosovitskiy2020image}. The reverse process of Mixup, \ie{} unmixing or decomposition, has been well-studied in low-level vision~\cite{han2021bid} and signal processing~\cite{hyvarinen2000independent}. 
CropMix mixes multiple cropped views of one image in a Mixup manner or CutMix manner to generate the mixed view as the input. Unlike Mixup, CropMix operates on different cropped views of a single image, hence no labeling changes are applied. CropMix aims to sample a richer input distribution for training and is complementary to Mixup. 

\textbf{Contrastive Learning}~\cite{hadsell2006dimensionality,he2020momentum,chen2020simple,chen2021exploring,Fang_surveyl} aims to pull positive pairs closer while pushing negatives apart in the representation space. Such contrastive pairs can be constructed by applying data augmentations. The multi-crop augmentation~\cite{swav} has been applied to contrastive learning methods~\cite{swav,wang2022asym,caron2021emerging, misra2020self}. However, the multi-crop augmentation forces the networks to handle multiple cropped views in every iteration, so that the computation overhead is greatly increased as a side effect. Also, for a certain resolution, multi-crop augmentation assigns the same crop scale to every cropping operation. In our study, we show that the key to effectiveness is to crop with distinct crop scales. The application scope of multi-crop is limited, it is effective for contrastive methods and clustering methods, but not for supervised methods~\cite{swav}. CropMix is more general and scales well to other vision tasks. 

\textbf{Masked Image Modeling} learns representations from reconstructing corrupted images. Autoencoders~\cite{bengio2009learning,hinton2006reducing} and denoising autoencoders~\cite{vincent2008extracting} are the pioneering work in this line. In the deep learning era, image colorization~\cite{zhang2016colorful} and image inpainting~\cite{pathak2016context} are adopted as pretext tasks for self-supervised learning.
Recently, motivated by the success of masked language modeling and vision transformers, masked image modeling methods~\cite{chen2020generative, bao2021beit, MaskedAutoencoders2021,dosovitskiy2020image} have achieved superior performance. 

\section{Method}

\begin{figure}[!htb]
     \centering
     \includegraphics[width = 13.8cm]
     {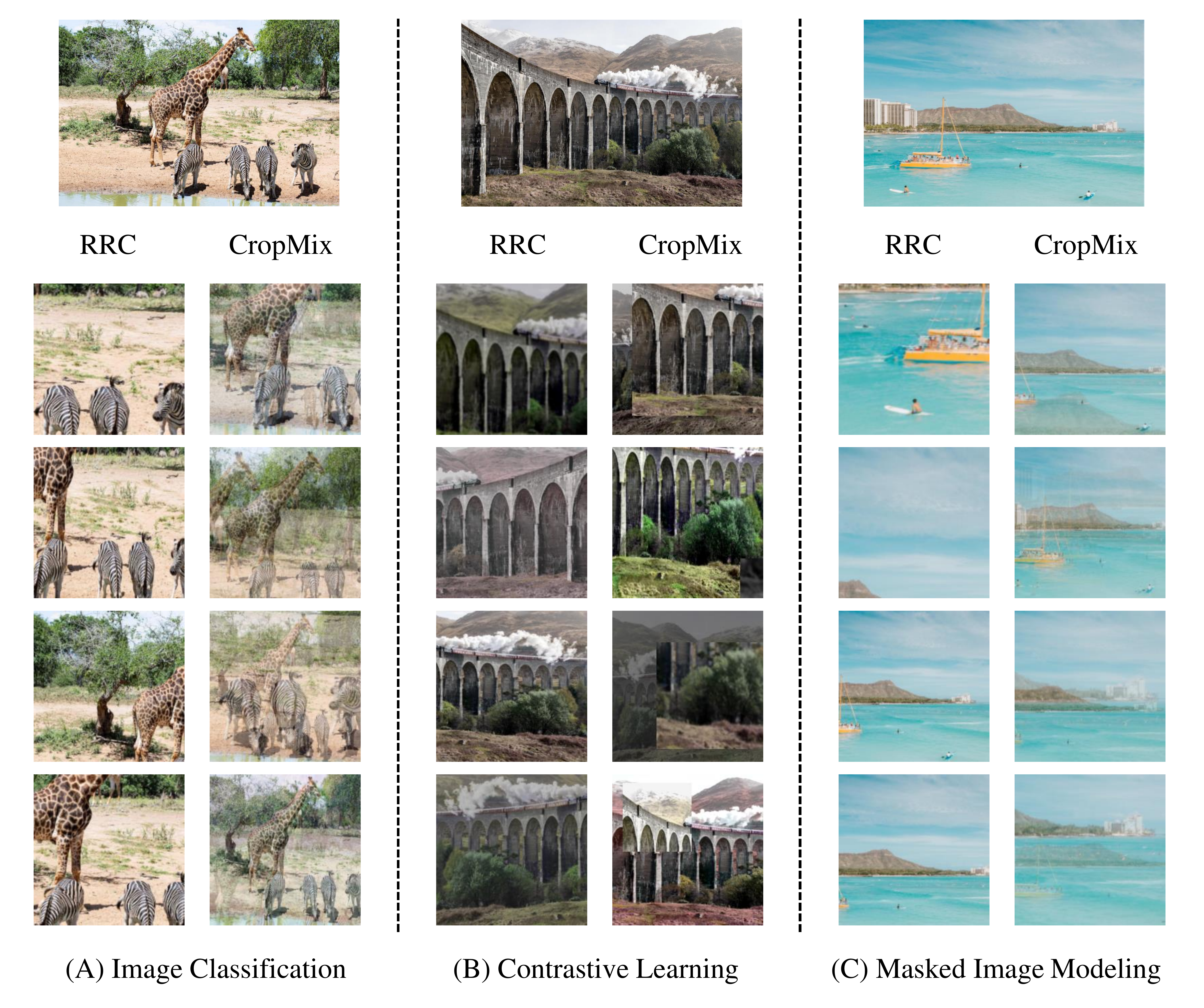}
     \caption{\textbf{Illustration of RRC (Random Resized Crop) and CropMix among 3 vision tasks}: image classification (A), contrastive learning (B), and masked image modeling (C). The mixing operations in CropMix are Mixup, CutMix, and Mixup for tasks A, B, and C, respectively. Original images are taken from~\cite{touvron2022deit}. Compared to RRC capturing single-scale information only, CropMix using the Mixup mixing operation combines the information from both a local fine-grained scale and a global coarse-grained scale. 
     Such multi-scale information benefits both image classification (A) and masked image modeling (C). For the contrastive learning (B), CropMix can construct better contrastive pairs for siamese representation learning.
     }
     \label{fig:compare_crop}
     \vspace{-0.5cm}
\end{figure}

\subsection{Random Resized Crop (RRC) Revisited}
Let $\Vec{X} \in \mathbb{R}^{C\times H \times W}$ denote an image, chosen from the original dataset $\mathcal{X}$, $c(\cdot)$ denote cropping operations, and $r(\cdot)$ denote resizing operations. By applying RRC, a cropped view $\Vec{X}'$ can be obtained as $\Vec{X}' = r(c(\Vec{X}))$. Cropping operations have two parameters, crop ratio and crop scale. Following the common practice, we fix the crop ratio as (3/4, 4/3), but with various crop scales. 

\subsection{CropMix}
\textbf{Cropping.} We define a set ${\mathcal{C}} = \{c_n\}_{n=1}^N$ containing $N$ cropping operations, where the wholistic crop scale is evenly divided to every cropping operation without overlapping. That is, for a crop scale range (0.01, 1.0) assigned to 3 cropping operations, the assigned crop scale ranges are (0.01, 0.34), (0.34, 0.67), and (0.67, 1.0), as shown in Figure~\ref{fig:ppline}. By doing so, we can guarantee that each cropping operation captures distinct information from different scales.
A set of $N$ cropped views are obtained via ${\mathcal{X}'} = \{r(c_n(\Vec{X}_n))\}_{n=1}^N$. 

\textbf{Mixing.} We denote mixing operations as $m(\cdot)$. We first randomly choose two cropped views from ${\mathcal{X}}'$ without replacement. $m(\cdot)$ takes these two cropped views as input and produces an intermediate mixing result. Then we further randomly choose one cropped view from ${\mathcal{X}}'$, also without replacement, and mix it with the intermediate mixing result to produce the next intermediate mixing result. By repeating the mixing procedure $N-1$ times, every $\Vec{X}'_i \in \mathcal{X}'$ contributes to the final mixed image.


For the mixing operation $m(\cdot)$, the mixing weight is randomly sampled from a beta distribution $Beta(\alpha, \alpha)$, where we set $\alpha$ as 0.4/$N$ for Mixup and 1.0 for CutMix. 
The algorithmic details of mixing operations are provided in the \textit{Appendix}.

\textbf{Intermediate augmentations.} Before every mixing operation, augmentations can be applied to the images involved in mixing. For classification tasks, we perform a channel permutation to the image with a smaller mixing weight. In contrastive learning, heavy intermediate augmentations are necessary to construct challenging contrastive pairs, whereas masked image modeling does not require any intermediate augmentations. 

\section{Classification}
In this section, we evaluate CropMix for its capability to train better image classifiers in multiple classification settings. 
We first study the effect of CropMix on CIFAR-10 and CIFAR-100~\cite{krizhevsky2009learning} datasets among \textbf{3} crop scale settings, \textbf{5} CNN models, and \textbf{2 }ViT models.

We then validate the performance of CropMix on ImageNet, where classification accuracy (generalization), calibration, robustness against adversarial attacks, corruption robustness, and robustness under distribution shifts are thoroughly evaluated. To verify the compatibility of CropMix with other regularizations and augmentations, we train multiple models on \textbf{3} training recipes. In terms of all evaluation metrics, CropMix consistently outperforms RRC among all models and training recipes.

\subsection{CIFAR}
\label{sec:cifar}
\begin{table*}[ht!]
  \centering
  \fontsize{8}{1}
  \selectfont
    \begin{tabular}{l|c|c|c|c|c|c}
    \toprule
     \Rows{Models}&\multicolumn{3}{c|}{CIFAR-10}&\multicolumn{3}{c}{CIFAR-100} \cr
    &(0.75, 1.0) & (0.25, 1.0) & (0.01, 1.0)  &(0.75, 1.0) &(0.25, 1.0) &(0.01, 1.0)\cr
    \midrule
    \gc{PreResNet18}& \gc{4.95}&\gc{5.24} & \gc{5.50}&\gc{24.48}&\gc{26.70} & \gc{26.86}\cr
    + CropMix&  5.04 & 4.80  & 4.92 &25.88  & 24.71 &  24.88\cr
    $\triangle$ & \re{+0.09} & \bl{-0.44} & \bl{-0.58} & \re{+1.00}  & \bl{-1.99} & \bl{-1.98}\cr
    \midrule
    \gc{Xception}& \gc{5.14}&\gc{5.18} & \gc{5.75}&\gc{24.68}&\gc{25.59} & \gc{27.23}\cr
    + CropMix&  5.27 & 4.99  & 4.69 & 25.69  & 24.98 & 24.60 \cr
    $\triangle$ & \re{+0.13} & \bl{-0.19} & \bl{-1.06} & \re{+1.01}  & \bl{-0.61} & \bl{-2.63}\cr 
    \midrule
    \gc{DenseNet121}& \gc{5.09}&\gc{5.44} & \gc{5.80}&\gc{24.52}&\gc{25.74} & \gc{27.01}\cr
    + CropMix&  5.22 & 4.64 & 4.76 & 25.49  & 24.35 & 24.16 \cr
    $\triangle$ & \re{+0.13} & \bl{-0.80} & \bl{-1.04} & \re{+0.97}  & \bl{-1.39} & \bl{-2.85}\cr 
    \midrule
    \gc{ResNeXt50}& \gc{5.38}&\gc{5.40} & \gc{5.57}&\gc{24.46}&\gc{26.70} & \gc{26.84}\cr
    + CropMix&  5.52& 4.50 & 4.83 & 25.45  & 24.94 &  24.63 \cr
    $\triangle$ & \re{+0.14} & \bl{-0.90} & \bl{-0.74} & \re{+0.99}  & \bl{-1.76} & \bl{-2.21}\cr 
    \midrule
    \gc{WRN-28-10}& \gc{5.31}&\gc{5.33} & \gc{5.65}&\gc{24.23}&\gc{26.02} & \gc{27.16}\cr
    + CropMix&  5.36 & 5.13  & 4.78  & 25.23 & 24.96 & 24.88 \cr
    $\triangle$ & \re{+0.05} & \bl{-0.20} & \bl{-0.87} & \re{+1.00}  & \bl{-1.06} & \bl{-2.28}\cr 
    \midrule    
    \gc{ViT}& \gc{5.40}&\gc{5.17} & \gc{5.79}&\gc{23.96}&\gc{26.22} & \gc{27.89}\cr
    + CropMix&  5.62 & 4.69  & 4.55 & 24.98 & 24.15 & 24.88 \cr
    $\triangle$ & \re{+0.22} & \bl{-0.48} & \bl{-1.24} & \re{+1.02}  & \bl{-2.07} & \bl{-3.01}\cr 
    \midrule    
    \gc{Swin}& \gc{5.08}&\gc{5.23} & \gc{5.81}&\gc{24.61}&\gc{25.89} & \gc{26.94}\cr
    + CropMix& 5.04 & 4.55  & 4.90 & 25.20 &24.92 & 24.67 \cr
    $\triangle$ & \bl{-0.04} & \bl{-0.68} & \bl{-0.91} & \re{+0.59}  & \bl{-0.97} & \bl{-2.27}\cr 
    \midrule    
    \gc{Average}& \gc{5.19}&\gc{5.28} & \gc{5.69}&\gc{24.47}&\gc{26.12} & \gc{27.13}\cr
    + CropMix& 5.29 & 4.75  & 4.77 & 25.41 &24.71 & 24.67 \cr    
    $\triangle$ & \re{+0.10} & \bl{-0.53} & \bl{-0.92} & \re{+0.94}  & \bl{-1.41} & \bl{-2.46}\cr
    \cr       
   \bottomrule
    \end{tabular}
     \caption{\textbf{Test top-1 error rate on CIFAR-10 and CIFAR-100 datasets. } We test \textbf{7} models on \textbf{3} crop scale settings. \bl{blue} indicates improved results. Average is the averaged result of \textbf{7} models and $\triangle$ is the difference between CropMix and RRC. Results are the mean of 5 independent trails. Standard deviations for CIFAR-10 and CIFAR-100 lie in (0.05, 0.16) and (0.08, 0.43), respectively. }
     \label{tab:cifar}
     \vspace{-0.5cm}
\end{table*}
\subsubsection{Task setting.}
We employ \textbf{5} CNN architectures (PreResNet18~\cite{he2016identity}, Xception~\cite{chollet2017xception}, DenseNet121~\cite{huang2017densely}, ResNeXt50~\cite{Xie2017}, WRN-28-10~\cite{Zagoruyko2016WRN}) and \textbf{2} ViT architectures (ViT~\cite{dosovitskiy2020image}, Swin~\cite{liu2021swin}). We set \textbf{3} crop scales: \textbf{(0.75, 1.0)}, \textbf{(0.25, 1.0)}, and \textbf{(0.01, 1.0)} for evaluation. Except for the crop scale, the training recipe follows YOCO~\cite{han2022yoco}, where models are trained for 300 epochs using the SGD optimizer and horizontal flip augmentation only in 32 $\times$ 32 resolution. For CropMix, we set the cropping operations as \textbf{2}, mixing operation as Mixup, and apply channel permutation as intermediate augmentation. 

\subsubsection{Results}
Table~\ref{tab:cifar} presents the results on both CIFAR-10 and CIFAR-100, where we report the \textit{best} results based the average of 5 independent trials. Among \textbf{7} models, CropMix significantly outperforms RRC when the range of the crop scale is wide. Surprisingly, with a range of (0.01, 1.0), the performance of CropMix even outperforms RRC with a range of (0.75, 1.0), in most of the cases.

However, for a narrow crop scale range of (0.75, 1.0), the cropped views obtained in CropMix are not distinct enough, failing to capture multi-scale information and hence unable to produce meaningful mixed images, which might be the reason for the performance drop. See Section~\ref{subsec:scale} for more analysis. 


\subsection{ImageNet}
\begin{table*}[!htbp]
  \centering
  \fontsize{8}{3}
  \selectfont
    \begin{tabular}{l|c|c|c|c|c|c|c|c|c}
    \toprule
     \Rows{Recipes}&\Rows{Models}& \multicolumn{1}{c|}{Generalization}&\multicolumn{1}{c|}{Calibration}&\multicolumn{1}{c|}{Attack}&\multicolumn{2}{c|}{Corruptions} &\multicolumn{3}{c}{Distribution shift} \cr
     & &Clean  & Clean, RMS$\downarrow$ &FGSM & Replace & Noise & IN-A & IN-R & IN-S \cr
     \midrule
     \gc{R1} & \gc{R-50} & \gc{76.59} & \gc{8.81} & \gc{21.01} & \gc{58.77} & \gc{72.92} & \gc{4.29} & \gc{35.30} & \gc{23.08} \cr
     +CropMix & R-50 & 77.60 & 7.73 & 23.94 & 59.29 & 73.82 & 6.53 & 39.12 & 25.56 \cr   
     $\triangle$ & R-50 &  \bl{+1.01}   &  \bl{-1.08} &  \bl{+2.93} &  \bl{+0.52} &  \bl{+0.90} &  \bl{+2.24} &  \bl{+3.82} &  \bl{+2.48} \cr
     \midrule
     \gc{R1} & \gc{R-200} & \gc{78.49} & \gc{11.99} & \gc{27.63} & \gc{59.87} & \gc{75.49} & \gc{10.84} & \gc{38.71} & \gc{27.05} \cr
     +CropMix & R-200 & 79.43 & 11.79 & 32.01 & 60.53 & 76.75 & 15.38 & 43.76 & 31.00 \cr 
     $\triangle$ & R-200 &   \bl{+0.94}   &  \bl{-0.20} &  \bl{+4.38} &  \bl{+0.66} &  \bl{+1.26} &  \bl{+4.54} &  \bl{+5.05} &  \bl{+3.95}  \cr     
     \midrule
     \gc{R1} & \gc{RepVGG} & \gc{75.28} & \gc{5.24} & \gc{8.50} & \gc{56.45} & \gc{71.74} & \gc{3.66} & \gc{35.69} & \gc{23.66} \cr
     +CropMix &RepVGG & 75.65 & 4.75 & 12.55 & 56.84 & 72.24 & 4.79 & 38.70 & 26.99 \cr 
     $\triangle$ & RepVGG &  \bl{+0.37}   &  \bl{-0.49} &  \bl{+4.05} &  \bl{+0.39} &  \bl{+0.50} & \bl{+1.13} &  \bl{+3.01} &  \bl{+3.33} \cr   
     \midrule
     \gc{R2} & \gc{R-50} & \gc{78.41} & \gc{30.62} & \gc{37.12} & \gc{60.12} & \gc{76.68} & \gc{9.67} & \gc{41.11} & \gc{28.22} \cr 
     +CropMix & R-50 & 78.71 & 19.62 & 37.47 & 61.87 & 77.19 & 12.43 & 44.25 & 32.39  \cr
     $\triangle$ & R-50 &  \bl{+0.30}   &  \bl{-11.00} &  \bl{+0.35} &  \bl{+1.75} &  \bl{+0.51} &  \bl{+2.76} &  \bl{+3.14} &  \bl{+4.17} \cr  
     \midrule
     \gc{R2} & \gc{RepVGG} & \gc{76.24} & \gc{24.72} & \gc{20.68} & \gc{59.77} & \gc{74.82} & \gc{6.53} & \gc{41.17} & \gc{26.40} \cr
     +CropMix &RepVGG & 76.39 & 21.66 & 21.46 & 60.30 & 74.95 & 7.75 & 42.99 & 28.23 \cr 
     $\triangle$ & RepVGG &  \bl{+0.15}   &  \bl{-3.06} &  \bl{+0.78} &  \bl{+0.53} &  \bl{+0.13} & \bl{+1.22} &  \bl{+1.82} &  \bl{+1.83} \cr   
     \midrule
     \gc{R3} & \gc{R-50} & \gc{78.67} & \gc{26.23} & \gc{34.97} & \gc{55.42} & \gc{76.95} & \gc{8.10} & \gc{40.80} & \gc{28.82} \cr
     +CropMix & R-50 & 78.82 & 26.00 & 36.02 & 55.62 & 77.07 & 9.68 & 41.96 & 29.36 \cr   
     $\triangle$ & R-50 &  \bl{+0.15}   &  \bl{-0.23} &  \bl{+1.05} &  \bl{+0.20} &  \bl{+0.12} &  \bl{+1.58} &  \bl{+1.16} &  \bl{+0.54} \cr      
     \midrule
     \gc{R3} & \gc{ConvNext} & \gc{81.72} & \gc{8.79} & \gc{33.02} & \gc{56.74} & \gc{79.87} & \gc{23.88} & \gc{46.68} & \gc{33.13} \cr 
     +CropMix & ConvNext & 81.92 & 8.70 & 38.86 & 57.03 & 80.26 & 25.65 & 48.29 & 34.84 \cr   
     $\triangle$ & ConvNext &  \bl{+0.20}   &  \bl{-0.09} &  \bl{+5.84} &  \bl{+0.29} &  \bl{+0.39} &  \bl{+1.77} &  \bl{+1.61} &  \bl{+1.71} \cr        
    \bottomrule
    \end{tabular}
     \caption{ \textbf{Results of applying CropMix on ImageNet across 3 training recipes}. Classification performance metrics includes generalization (top-1 accuracy), calibration (RMS error), robustness against adversarial attacks (top-1 accuracy), corruption robustness (top-1 accuracy), and robustness under distribution shifts (top-1 accuracy). Improved results are highlighted in \bl{blue} color. $\triangle$ is the difference between CropMix and RRC.
     }
     \label{tab:imagenet}
          \vspace{-0.5cm}
\end{table*}
\subsubsection{Experimental setup}
We adopt \textbf{3} training recipes (R1, R2, R3), from simple to complex, to thoroughly validate the performance of CropMix. R1 is a standard ImageNet training recipe taken from CutMix~\cite{yun2019cutmix} whereas R3 is a complex one used in training ConvNext~\cite{liu2022convnet}. The complexity of R2 is between R1 and R3, please refer to \textit{Appendix} for training details. We take ResNet-50~\cite{resnet}, ResNet-200~\cite{resnet}, RepVGG-A2~\cite{ding2021repvgg}, and ConvNext-T~\cite{liu2022convnet} as our models. We choose an aggressive crop scale range, (0.01, 1.0) for CropMix while keeping the range as (0.08, 1.0) for RRC. The CropMix setting is as follows: the number of cropping operations are randomly chosen from (2, 3, 4), the mixing operation is Mixup, and channel permutation is adopted as intermediate augmentation. 

\subsubsection{Evaluation protocol}
Following the evaluation protocol used in YOCO~\cite{han2022yoco}, other than classification accuracy, we also thoroughly evaluate calibration error, robustness against adversarial attacks, corruption robustness, and robustness under distribution shifts.

\textbf{Calibration.} We follow the evaluation protocol used in PixMix~\cite{hendrycks2021pixmix}, where RMS calibration error~\cite{hendrycks2018deep} and adaptive binning~\cite{nguyen2015posterior} are applied. 

\textbf{Adversarial attack.} We employ FGSM~\cite{fgsm} attack with a budget of 8/255 to verify the robustness against adversarial attacks. 

\textbf{Corruption robustness.} To verify the generalization ability and robustness of image classifiers to unseen environments, we borrow two challenging corrupted test sets from Co-Mixup~\cite{kim2021co}. The two test sets, (1) Random background replacement and (2) Gaussian noise, are crafted by replacing the background with a random image and adding Gaussian noise to the background, correspondingly.

\textbf{Distribution shift.} When the train and test distributions are not \textit{i.i.d}, which is frequent in practice, the classification accuracy can plummet. Hence improving the robustness to unforeseen data shifts is important to real-world applications. We employ ImageNet-Adversarial~\cite{hendrycks2021natural}, ImageNet-Rendition~\cite{hendrycks2021many}, and ImageNet-Sketch~\cite{wang2019learning}, \textbf{3} commonly used yet challenging ImageNet test sets, to verify the performance of image classifiers against input data distribution shifts.  

\subsubsection{Results}
Quantitative results are summarized in Table~\ref{tab:imagenet} while qualitative results and visualizations are presented in the \textit{Appendix}. Reported results are the \textit{last} results. Regardless of training recipes and models, CropMix improves all classification performance metrics across-the-board, showing its generality and superiority. Notably, CropMix is far better than RRC under distribution shift, where the improvement margin in top-1 accuracy can be up to 5.05\%. The input distribution sampled via CropMix is richer than the one sampled via RRC, such an input distribution might compensate distribution shift and can better adapt to unseen environments.

It is also notable that when more augmentations and regularizations are applied, the improvement margin of CropMix tends to decrease, showing that CropMix shares some common regularization effects with other regularizations. However, when up against data distribution shifts, CropMix still brings a considerable performance gain. This suggests how to sample the input distribution can play a crucial role in improving robustness under distribution shifts. 

\section{Contrastive Learning}
\label{sec:contrastive}
\begin{table}[ht!]
  \centering
  \fontsize{6.3}{6}
  \selectfont
    \begin{tabular}{lc|c|c|c|c|c|c|c|c|c|c|cc}
    \toprule
    \Rows{Method}&\multicolumn{3}{c|}{ImageNet classification} &\multicolumn{3}{c|}{VOC detection} &\multicolumn{3}{c|}{COCO detection}  &\multicolumn{3}{c}{COCO instance seg} \cr
    & linear probe&  10\% label & 1\% label  & AP$_\text{50}$ & AP & AP$_\text{75}$ & AP$_\text{50}$ & AP & AP$_\text{75}$ & AP$^\text{mask}_\text{50}$ & AP$^\text{mask}$ & AP$^\text{mask}_\text{75}$   \cr
    \midrule
    \gc{MoCo v2} & \gc{65.8}  & \gc{57.3} & \gc{14.6} & \gc{81.4} & \gc{55.8} & \gc{62.5}  & \gc{56.3}  & \gc{37.2} & \gc{40.2} & \gc{53.2} & \gc{32.8} & \gc{34.9}   \cr
    + ScaleMix & 67.2  & 58.3  & 14.6 & 81.8 & 55.7 & 61.0 & 56.2 & 37.2 & 40.1 & 53.1 & 32.7 & 34.8 \cr    
    + CropMix & 67.8  & 58.9  & 18.7 & 81.9 & 56.3 & 62.4 & 56.2 & 37.0 & 40.1 & 53.0 & 32.7 & 34.7 \cr
    \bottomrule
    \\
    \end{tabular}
     \caption{\textbf{Results of contrastive learning}. MoCo v2 here is an improved version of vanilla MoCo v2. The evaluation metric of ImageNet classification is Top-1 accuracy.  }
     \label{tab:contrastive}
     \vspace{-0.5cm}
\end{table}

A recent study~\cite{wang2022asym} shows that the asymmetry design in the siamese network can achieve superior performance in contrastive learning.  ScaleMix~\cite{wang2022asym} is one of the powerful design choices in the asymmetry representation learning. ScaleMix can be treated as a special case of CropMix, where the setting is $N = 2$, mixing in a CutMix manner, but two views are obtained via same-scale cropping. We show that, with a little change, from same-scale cropping to multi-scale cropping, the learned embedding can yield a considerable performance gain in downstream classification tasks. 

\subsection{Experimental setup}
We follow the pre-training setting in~\cite{wang2022asym}, where we train an improved version of MoCo v2 for 100 epochs. This version adds a fully connected layer (2048 dimension, with BatchNorm) before the projection head (2-layer MLP). Downstream tasks follow the setting used in YOCO~\cite{han2022yoco}, details are available in the \textit{Appendix}.

\subsection{Results}
Table~\ref{tab:contrastive} depicts the results, where results reported in semi-supervised fine-tuning and VOC detection are the average of \textbf{3} independent runs. Among classification tasks, the improvement margins of CropMix over baseline are 2.0\%, 2.6\%, and 4.1\% for respective setting. Compared to ScaleMix, mixing cropped views obtained with distinct crop scales brings significant gain under the semi-supervised setting, where the 1\% label fine-tuning are improved by 4.1\%, showing the superior transferability of CropMix to small-scale datasets. 
However, when learned features are transferred to detection and segmentation tasks, there is no trend showing that either ScaleMix or CropMix helps.

\section{Masked Image Modeling}
CropMix can sample a richer input distribution from the dataset distribution, this section aims to explore whether learning to reconstruct such CropMix produced images can be beneficial to the pre-training of masked image modeling. 

\begin{wraptable}{r}{8.5cm}
\vspace{-.2em}
  \fontsize{9}{1}
  \selectfont
    \begin{tabular}{l|c|c|c|c|c}
    \toprule
     \Rows{Methods}&\Rows{Data}& \multicolumn{1}{c|}{Generalization}&\multicolumn{3}{c}{Distribution shift} \cr
     & &Clean & IN-A & IN-R & IN-S \cr
     \midrule
     MAE & 100 \% & 82.4 & 28.7 & 48.3 & 33.6 \cr
     +CropMix & 100 \%  & 82.3 & 28.4 & 48.6 & 34.4 \cr
     \midrule
     MAE & 10 \%  & 51.9 & 2.0 & 19.7 & 8.3\cr
     +CropMix & 10 \% & 52.6 & 2.1 & 20.5 & 8.9 \cr     
    \bottomrule
    \end{tabular}
\caption{\textbf{Results on masked image modeling}. We pre-train and fine-tune MAE models (ViT-base backbone) on 10\% and 100\% ImageNet data. CropMix is applied to the pre-training stage only. The evaluation metric is Top-1 accuracy. Overall, CropMix can form a better learning objective when dataset size is restricted.}\label{tab:mim}
\vspace{-1cm}
\end{wraptable} 


\subsection{Experimental setup}
We choose MAE~\cite{MaskedAutoencoders2021} with ViT -base~\cite{dosovitskiy2020image} backbone as our baseline. All models are pre-trained for 200 epochs and 300 epochs for 100\% and 10\% ImageNet data, respectively. In the fine-tuning stage, we assign corresponding label fractions to pre-trained models, and models are fine-tuned for 100 epochs. Details are presented in the \textit{ Appendix}.

\subsection{Results}
Results are depicted in Table~\ref{tab:mim}. We measure test-time accuracy and robustness under distribution shifts. Applying CropMix is beneficial to all evaluation metrics on small-scale datasets, while also showing competitive results on large-scale datasets. 

\section{Analysis and Discussion}
Here we aim to explore the optimal setting of CropMix, verify multiple design choices, and find the application scope. All classification experiments are performed on the 10\% ImageNet dataset with the ResNet-50 model. Reported results are the average of \textbf{2} independent trials.

\textbf{Number of crops.} By default, we randomly choose $N$ from (2,3,4) for ImageNet classification. Here, we set $N$ as 2, 3, 4, 5 and also randomly choose it from (2,3) and (2,3,4,5). 
CropMix is flexible with the number of cropped views $N$, as studied in Table~\ref{tab:num_crops}. Mixing with more crops can boost the performance, but the improvement tends to be marginal when $N > 4$. Randomly choosing $N$ is more efficient than fixing $N$ in both performance and computational efficiency.

\textbf{Single-scale or multi-scale?}
We assign the same crop scale to all cropping operations in CropMix, referred to as single-scale. Table~\ref{tab:same_scale} studies this design across different number of crops. The multi-scale crop always outperforms single-scale crop design by considerable margins, especially when more crops are used. We observe a similar trend in contrastive learning experiments as presented in Section~\ref{sec:contrastive}. Multi-scale cropping allows us to capture information of distinct scales and avoid capturing overlapping information. Such a multi-scale cropping design benefits visual recognition in general. 

\textbf{Resizing resolution.}
We compare different resizing resolutions in Table~\ref{tab:resize}, where both train and test resolutions are adjusted. Interestingly, CropMix benefits more when the resolution is higher. As studied in~\cite{touvron2019fixing}, the crop size can affect the activation statistics, where models trained with higher resizing resolution have larger activation maps. We hypothesise that larger activation maps can better leverage the multi-scale information from the mixed image created by CropMix. 

\begin{table*}[t]
\vspace{-.2em}
\centering
\subfloat[
\textbf{Number of crops}.
\label{tab:num_crops}
]{
\begin{minipage}{0.28\linewidth}{\begin{center}
\tablestyle{4pt}{1.05}
\begin{tabular}{x{30}x{24}}
Num & Acc \\
\shline
1 & 52.57  \\
2 & 54.89  \\
3 &  54.94 \\
4 & 55.11  \\
5 &  55.14 \\
(2,3) & 55.06 \\
(2,3,4) & \baseline{55.15} \\
\end{tabular}
\end{center}}\end{minipage}
}
\hspace{2em}
\subfloat[
\textbf{Single-scale or multi-scale?}
\label{tab:same_scale}
]{
\begin{minipage}{0.28\linewidth}{\begin{center}
\tablestyle{4pt}{1.05}
\begin{tabular}{x{30}x{24}x{24}}
Num & Single & Multi \\
\shline
1 & 52.57 & -  \\
2 & 54.74 & 54.89   \\
3 &  54.76 &54.94  \\
4 &  54.53& 55.11   \\
5 &  54.49& 55.14  \\
(2,3) & 54.60& 55.06  \\
(2,3,4) & 54.52&\baseline{55.15}   \\
\end{tabular}
\end{center}}\end{minipage}
}
\hspace{2em}
\subfloat[
\textbf{Resizing resolution}. 
\label{tab:resize}
]{
\begin{minipage}{0.28\linewidth}{\begin{center}
\tablestyle{4pt}{1.05}
\begin{tabular}{x{24}x{24}x{30}}
Res & RRC & CropMix \\
\shline
32 &  18.64 & 19.62 \\
64 &  32.08 & 33.95 \\
128 &  42.32 & 45.44 \\
224 &  52.89 & \baseline{55.15} \\
384 &  49.00 & 52.48 \\
\multicolumn{3}{c}{~}\\
\multicolumn{3}{c}{~}\\
\end{tabular}
\end{center}}\end{minipage}
}
\hspace{2em}
\subfloat[
\textbf{Interpolation mode}. 
\label{tab:inter}
]{
\begin{minipage}{0.28\linewidth}{\begin{center}
\tablestyle{4pt}{1.05}
\begin{tabular}{x{24}x{24}x{30}}
Mode & RRC & CropMix \\
\shline
Nearest & 51.45 & 54.07 \\
Bilinear & 52.89 & \baseline{55.15} \\
Bicubic & 52.15 & 55.05 \\
\multicolumn{3}{c}{~}\\
\multicolumn{3}{c}{~}\\
\end{tabular}
\end{center}}\end{minipage}
}
\hspace{2em}
\subfloat[
\textbf{Crop scale}. 
\label{tab:crop_scale}
]{
\begin{minipage}{0.28\linewidth}{\begin{center}
\tablestyle{4pt}{1.05}
\begin{tabular}{x{36}x{24}x{30}}
Range & RRC & CropMix \\
\shline
(0.01, 1) & 52.57 & \baseline{55.15}  \\
(0.08, 1) & 52.89 & 54.54  \\
(0.4, 1) & 50.93 & 52.13  \\
(0.6, 1) & 49.36 & 49.94  \\
(0.8, 1) & 46.22 & 46.03 \\
\end{tabular}
\end{center}}\end{minipage}
}
\hspace{2em}
\subfloat[
\textbf{Intermediate aug}.
\label{tab:aug}
]{
\begin{minipage}{0.28\linewidth}{\begin{center}
\tablestyle{4pt}{1.05}
\begin{tabular}{x{30}x{18}x{18}x{18}}
Augs & Before & After & Both  \\
\shline
H flip &  53.87 &  53.99 & 53.97 \\
V flip &  54.26 &  51.19 & 50.69 \\
Permute & \baseline{55.15} &  54.10 & 54.34  \\
Color jit & 54.39 & 55.36 & 55.82  \\
AutoAug & 54.51  & 57.60 & 57.62  \\
\end{tabular}
\end{center}}\end{minipage}
}
\hspace{2em}
\vspace{-.1em}
\caption{\textbf{CropMix ablation experiments} with ResNet-50 model on 10 \% ImageNet. We report the top-1 classification accuracy. Default settings are marked in \colorbox{baselinecolor}{gray}. 
}
\label{tab:ablations}
\vspace{-0.3cm}
\end{table*}

\textbf{Interpolation mode.}
We study the effect of interpolation mode in Table~\ref{tab:inter}. We change the interpolation mode in both the training and testing stages. Surprisingly, CropMix tends to bring more performance gain when the interpolation mode is complex. A more complex multivariate interpolation mode can produce smoother results with fewer interpolation artifacts, which might help to preserve the multi-scale information, and thus adapts better to CropMix. 

\textbf{Impact of crop scale.}
\label{subsec:scale}
Table~\ref{tab:crop_scale} studies the influence of crop scale. CropMix is powerful with a wider range of crop scale while being ineffective with a narrow crop scale. More distinctive cropped views are necessary for the superior performance of CropMix to be realised. The trends also hold for CIFAR classification tasks studied in Section~\ref{sec:cifar}. Benefiting from the property of CropMix suppressing labeling errors, CropMix can be applied to various datasets with an aggressive crop scale range, \ie{}, (0.01, 1.0), without the need to performing grid search.

\textbf{Necessity of intermediate augmentations.}
In Table~\ref{tab:aug} we show the effect of applying intermediate augmentations. Here \textit{before} and \textit{after} means augment before mixing and after mixing, where\textit{ before} can be treated as applying intermediate augmentations while \textit{after} is the normal procedure of performing augmentations. Photometric transformation as intermediate augmentations tends to be beneficial for classification tasks whereas geometric transformation is not that effective, as the cropping step already involves geometric transformation.

More results and analysis, including \textbf{Choice of mixing operations} and \textbf{Sensitivity of mixing weights} are provided in the \textit{Appendix}.

\section{Limitation}
\label{sec:limitation}
CropMix shows its superiority and generality in multiple vision tasks empirically, however, we do not develop the theoretical foundation of CropMix, which limits the interpretability. 

\section{Conclusion}
We propose CropMix as a data processing method, where positive results in a variety of computer vision tasks and datasets are obtained by applying CropMix. Though cropping is an essential data processing step, it has been rarely studied in previous research. We hope our study will inspire more research work in exploiting the importance of sampling the input distribution. 

\section{Acknowledgement}
We thank Pengfei Fang for proofreading the paper. 

\bibliographystyle{abbrv} 
\bibliography{egbib}
\newpage

\appendix
\section{Method}
Given two same sized images, denoted by $X$ and $Y$, the combining operations for Mixup~\cite{zhang2017mixup} and CutMix~\cite{yun2019cutmix} to produce the mixed image $Z$ are as follows:

\textbf{Mixup.}
$Z = \lambda \cdot X + (1- \lambda) \cdot Y $, where the mixing weight (combination ratio) $\lambda$ is randomly sampled from a beta distribution $Beta(\alpha, \alpha)$, for $\alpha \in(0, \infty)$.

\textbf{CutMix.}
$Z = M \odot X + (1 - M) \odot Y$, where $\odot$ denotes element-wise multiplication and $M{\in}\{0,1\}^{H{\times}W}$ denotes a binary mask.
$M$ is generated via sampling the coordinates of a bounding box, $B{=}\left(a,b,w,h\right)$, where $(a,b)$ is the center of the box and $(w,h)$ is the box size. $a, b$ are randomly sampled from $(0, W)$ and $(0, H)$, while $w, h$ are sampled following $w{=}W\sqrt{\lambda}$, $h{=}H\sqrt{\lambda}$, where $\lambda$ is sampled from a beta distribution $Beta(\alpha, \alpha)$. 


\begin{wraptable}{r}{8.5cm}
\vspace{-.2em}
    \begin{tabular}{c|ccc}
    \toprule
    Procedure& R1 & R2   & R3  \cr
    \midrule
    Train Res & 224 & 224 & 224 \cr
    Test Res & 224 & 224 & 224  \cr
    \midrule
    Epochs & 300& 300 & 300 \cr
    \midrule
    Batch size & 256 & 256 & 4096 \cr
    Optimizer & SGD-M & SGD-M & AdamW  \cr
    LR & 0.1 & 0.1 &  4e-3  \cr
    LR decay& step & cosine &cosine \cr
    decay rate & 0.1 & - & - \cr
    decay epochs & 75 & - & -  \cr
    Weight decay 1e-4 & 2e-4 & 5e-2 \cr
    Warmup epochs & 0 & 5 & 20\cr
    \midrule
    Label smoothing $\varepsilon$ & - & 0.1 & 0.1 \cr
    \midrule
    Horizontal flip& \cmark & \cmark & \cmark \cr
    Rand Augment & \xmarkg & \xmarkg & \cmark  \cr
    Trivial Augment & \xmarkg & \cmark & \xmarkg \cr
    ColorJitter & \xmarkg & \xmarkg & \cmark   \cr
    Mixup alpha  & - & 1.0 & 0.8  \cr
    Cutmix alpha & - & - &1.0\cr
    Erasing prob& - & 0.1 & 0.25\cr
    \midrule
    EMA & \xmarkg & \xmarkg & \cmark  \cr
    \midrule
    Test  crop ratio & 0.875 & 0.875 & 0.875 \cr
    \bottomrule
    \end{tabular}
    \caption{Summary of our training recipes on ImageNet. R1 is taken from CutMix~\cite{yun2019cutmix} while R3 is adopted from ConvNext~\cite{liu2022convnet}. }
    \label{tab:recipe}
\vspace{-1cm}
\end{wraptable} 

\section{Implementation Details}
\subsection{ImageNet Classification}
We present the details of \textbf{3} training recipes in Table~\ref{tab:recipe}. All models are trained using 8 GPUs. Recipe 3 in default employs 32 GPUs in parallel for training. However, we only use 8 GPUs, each with a batch size of 64, and 8 gradient accumulation steps, due to our limited computation resources. It is also known that through the effective batch size is identical (8 * 64 * 8 = 4096), it may bring performance gap after applying gradient accumulation/. For a fair comparison, all reported results are based on our reproduced results.

\subsection{Contrastive Learning}
\subsubsection{Pre-training}
In the pre-training stage, we follow the training setting of Asym-siam~\cite{wang2022asym}. An improved version of MoCo v2 is trained for 100 epochs with SGD optimizer. Weight decay and momentum are set as 0.0001 and 0.9, respectively. The batch size is 256 and is distributed to 4 GPUs. The initial learning rate is 0.03 with a cosine decay schedule. The crop scale is set to (0.2, 1.0) for both RRC and CropMix. 

\subsubsection{Linear probe}
We follow the linear probe setting used in Simsiam~\cite{chen2021exploring}. We employ LARS optimizer~\cite{you2017large}. The linear classifier layer is trained for 90 epochs with an initial learning rate of 0.2. The learning rate is adjusted with a cosine decay schedule. The total batch size is 4096 and is distributed over 8 GPUs. Noted we do not apply any weight decay in linear probe stage. 

\subsubsection{Fine-tuning}
We adopt the semi-supervised fine-tuning setting from Jigsaw Clustering~\cite{chen2021jigsaw}. 

Regardless of label fractions, pre-trained models are fine-tuned for 100 epochs using SGD optimizer with a momentum of 0.9. The batch size is 256 and is distributed over 4 Gpus. No weight decay is applied. Learning rate is decayed by a factor of 0.1 at 30, 60, and 90 epochs.

For 10\% label fraction, the trunk learning rate is 0.01 while the last layer learning rate is 0.2. For 1\% label fraction, the learning rate for the trunk is set as 0.02 and is set as 5 for the final layer. 

\subsubsection{VOC detection. }
We identically follow the setting in YOCO~\cite{han2022yoco}. Faster R-CNN~\cite{ren2015faster} with the C4-backbone is fine-tuned on VOC 2007 trainval + 2012 train and evaluated on the VOC 2007 test. The batch size is 8 and models are trained on 4 GPUs. The total iteration is 48K and the base learning rate is 0.01. All reported results are the average over 3 trials under. Results of ScaleMix and baseline are reproduced by us using the corresponding official pre-trained ResNet-50 models.

\textbf{COCO detection and instance segmentation.}
We use Mask R-CNN~\cite{he2017mask} with a C4 backbone as our model. Training settings are similar to the one used in VOC detection except for total iterations of 18K. Results of ScaleMix and baseline are also reproduced by us under this setting.

\subsection{Masked Image Modeling}
\subsubsection{Pre-training}
We employ MAE~\cite{MaskedAutoencoders2021} with ViT-base~\cite{dosovitskiy2020image} backbone as our baseline. Models are pre-trained for 300 epochs and 200 epochs for 10\% images and 100\% images, respectively. The warmup epochs are 20. We use AdamW optimizer with a base learning rate of 0.00015. The learning rate scheduler is a cosine decay schedule. Models are trained on 4 GPUs, each with a batch size of 64, and gradients are accumulated for 16 steps. So the effective batch size is 4 * 64 * 16 = 4096. The same crop scale, (0.2, 1.0), is applied to both RRC and CropMix. All reported results are our reproduced results using the official code. 

 \subsubsection{Fine-tuning}
Our fine-tuning setting is identical to MAE~\cite{MaskedAutoencoders2021}. Pre-trained weights are fine-tuned for 100 epochs with 5 warmup epochs. The effective batch size is 1024 ( 4 * 32 * 8), where models are trained on 4 GPUs, each with a batch size of 32, and 8 gradient accumulation steps. Random resized crop, horizontal flip, Mixup~\cite{zhang2017mixup}, CutMix~\cite{yun2019cutmix}, and random erasing~\cite{zhong2020random} are employed as data augmentation. The base learning rate is 0.0005 and is scheduled with a cosine scheduler. For 10\% and 100\% data, the fine-tuning setting is identical except for dataset size.

\section{Additional analysis}
\subsection{Choice of Mixing Operations}
We individually study the choice of mixing operations on image classification, contrastive learning, and masked image modeling. All settings follow  the same setting used in the main paper. 

\textbf{Image classification.}
We employ ResNet-50 model and training recipe R1 to study the choice of mixing operations on image classification tasks. We report both 10\% and 100\% ImageNet results in Table~\ref{tab:imagenet_supp}. When the dataset size is small, mixing in a Mixup manner can achieve superior performance in all evaluation metrics. For the full-size ImageNet, CutMix yields a high test top-1 accuracy (77.94 \%) while its performance is inferior to the Mixup in the distribution shift evaluation. 

\begin{table*}[!htbp]
  \centering
  \fontsize{8}{3}
  \selectfont
    \begin{tabular}{l|c|c|c|c|c|c|c|c|c}
    \toprule
     \Rows{Operations}&\Rows{Data}& \multicolumn{1}{c|}{Generalization}&\multicolumn{1}{c|}{Calibration}&\multicolumn{1}{c|}{Attack}&\multicolumn{2}{c|}{Corruptions} &\multicolumn{3}{c}{Distribution shift} \cr
     & &Clean  & Clean, RMS$\downarrow$ &FGSM & Replace & Noise & IN-A & IN-R & IN-S \cr
     \midrule
     \gc{Baseline} & \gc{100\%} & \gc{76.59} & \gc{8.81} & \gc{21.01} & \gc{58.77} & \gc{72.92} & \gc{4.29} & \gc{35.30} & \gc{23.08} \cr
     +Mixup & 100\% & 77.60 & 7.73 & 23.94 & 59.29 & 73.82 & 6.53 & 39.12 & 25.56 \cr   
     +CutMix & 100\% & 77.94 & 7.19 & 25.03 & 58.05 & 74.20 & 5.97 & 36.98 & 24.35 \cr  
     \midrule
     \gc{Baseline} & \gc{10\%} & \gc{52.57} & \gc{18.98} & \gc{7.59} & \gc{36.31} & \gc{44.28} & \gc{1.49} & \gc{18.60} & \gc{7.58} \cr
     +Mixup & 10\% & 55.15 & 15.91 & 4.86 & 38.98 & 47.45 & 1.56 & 20.95 & 8.64 \cr   
     +CutMix & 10\% & 53.48 & 17.48 & 8.60 & 37.89 & 42.64 & 1.51 & 17.59 & 7.30 \cr  
     \midrule     

    \end{tabular}
     \caption{ \textbf{Choose of mixing operations on image classification}. We employ classification performance metrics including generalization (top-1 accuracy), calibration (RMS error), robustness against adversarial attacks (top-1 accuracy), corruption robustness (top-1 accuracy), and robustness under distribution shifts (top-1 accuracy).
     }
     \label{tab:imagenet_supp}
\end{table*}

\textbf{Contrastive learning.}
We present our results in Table~\ref{tab:contrastive_supp}. CropMix with Mixup as mixing operation performs poorly, with only limited improvement margins compared to baseline. One possible explanation is that: the objective of contrastive pre-training is instance discrimination, CropMix with Mixup aggregates the multi-scale information to a single image at the pixel level, hence decreasing the difficulty of the learning objective and failing to produce a general visual representation.  
\begin{table}[ht!]
  \centering
  \fontsize{6.3}{6}
  \selectfont
    \begin{tabular}{lc|c|c|c|c|c|c|c|c|c|c|cc}
    \toprule
    \Rows{Operations}&\multicolumn{3}{c|}{ImageNet classification} &\multicolumn{3}{c|}{VOC detection} &\multicolumn{3}{c|}{COCO detection}  &\multicolumn{3}{c}{COCO instance seg} \cr
    & linear probe&  10\% label & 1\% label  & AP$_\text{50}$ & AP & AP$_\text{75}$ & AP$_\text{50}$ & AP & AP$_\text{75}$ & AP$^\text{mask}_\text{50}$ & AP$^\text{mask}$ & AP$^\text{mask}_\text{75}$   \cr
    \midrule
    \gc{MoCo v2} & \gc{65.8}  & \gc{57.3} & \gc{14.6} & \gc{81.4} & \gc{55.8} & \gc{62.5}  & \gc{56.3}  & \gc{37.2} & \gc{40.2} & \gc{53.2} & \gc{32.8} & \gc{34.9}   \cr
    + Mixup & 67.8  & 58.9  & 18.7 & 81.9 & 56.3 & 62.4 & 56.2 & 37.0 & 40.1 & 53.0 & 32.7 & 34.7 \cr
     + CutMix & 65.8  & 57.7  & 16.8 &81.7 & 55.5 & 62.3 & 56.4 & 37.1 & 39.9 & 53.3 & 32.6 & 34.7 \cr   
    \bottomrule
    \\
    \end{tabular}
     \caption{\textbf{Choose of mixing operations on contrastive learning}. MoCo v2 here is an improved asymmetric version of vanilla MoCo v2. The evaluation metric of ImageNet classification is Top-1 accuracy.  }
     \label{tab:contrastive_supp}
\end{table}

\textbf{Masked Image Modeling.}
Results are presented in Table~\ref{tab:mim_sup}. Interestingly, similar to the results presented in image classification tasks, the performance of applying Mixup and CutMix also varies with dataset size, where Mixup suits small-scale datasets while CutMix needs more data for the superior performance to be realized. 
\begin{table*}[ht!]
  \centering
  \fontsize{8}{1}
  \selectfont
    \begin{tabular}{l|c|c|c|c|c}
    \toprule
     \Rows{Operations}&\Rows{Data}& \multicolumn{1}{c|}{Generalization}&\multicolumn{3}{c}{Distribution shift} \cr
     & &Clean & IN-A & IN-R & IN-S \cr
     \midrule
     MAE & 100 \% & 82.4 & 28.7 & 48.3 & 33.6 \cr
     +Mixup & 100 \%  & 82.3 & 28.4 & 48.6 & 34.4 \cr
     +CutMix & 100 \%  & 82.5 & 29.9 & 49.3 & 34.9 \cr
     \midrule
     MAE & 10 \%  & 51.9 & 2.0 & 19.7 & 8.3\cr
     +Mixup & 10 \% & 52.6 & 2.1 & 20.5 & 8.9 \cr   
     +CutMix & 10 \%  & 51.8 & 2.2 & 19.8 & 8.0 \cr
    \bottomrule
    \end{tabular}
\caption{\textbf{Choose of mixing operations on masked image modeling}. We pre-train and fine-tune MAE models (ViT-base backbone) on 10\% and 100\% ImageNet data. CropMix is applied to the pre-training stage only, in either Mixup manner or CutMix manner. The evaluation metric is Top-1 accuracy. Overall, mixing with Mixup behaves better when data is limited, while mixing with CutMix tends to be helpful when dataset is large.}
\label{tab:mim_sup}
\end{table*} 

\subsection{Sensitivity of Mixing Weights}
\begin{figure*}[!htb]
     \centering
     \includegraphics[width = 7cm]
     {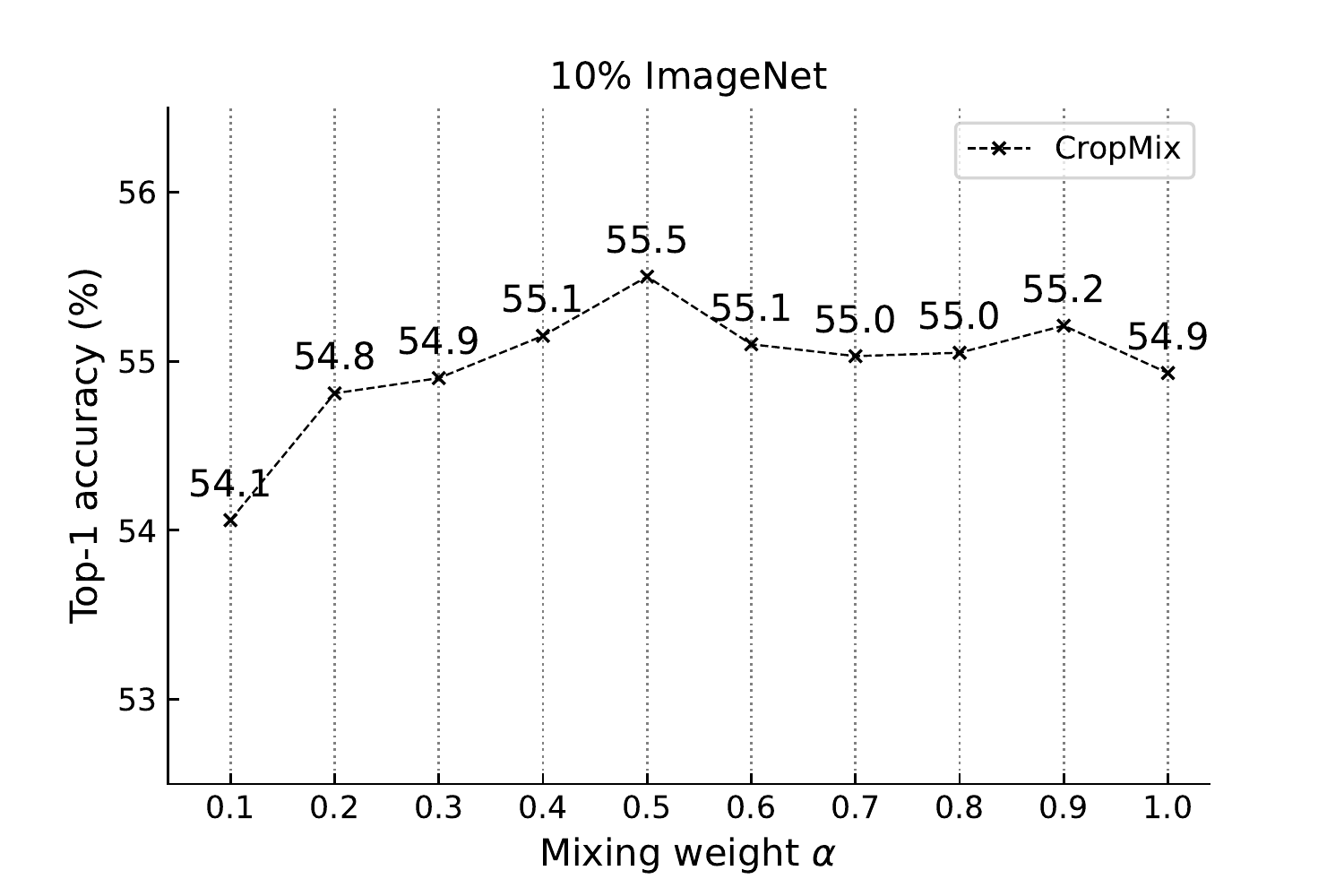}
     \caption{\textbf{Sensitivity of mixing weights}. Test top-1 accuracy on 10\% ImageNet is reported. We randomly sample the mixing weight from a beta distribution $\mathrm{Beta}(\alpha/N, \alpha/N)$. }
     \label{fig:weight}
\end{figure*}

Figure~\ref{fig:weight} presents our study on the sensitivity analysis of mixing weights. We sample the mixing weight from a Beta distribution $Beta(\alpha/N, \alpha/N)$, where $N$ is the number of mixing operations. When $\alpha > 0.1$, the performance of CropMix varies with the mixing weight slightly, where  $\alpha = 0.5$ yields the best performance. 

\section{Visualization}

\begin{figure*}[!htb]
     \centering
     \includegraphics[width = 14cm]
     {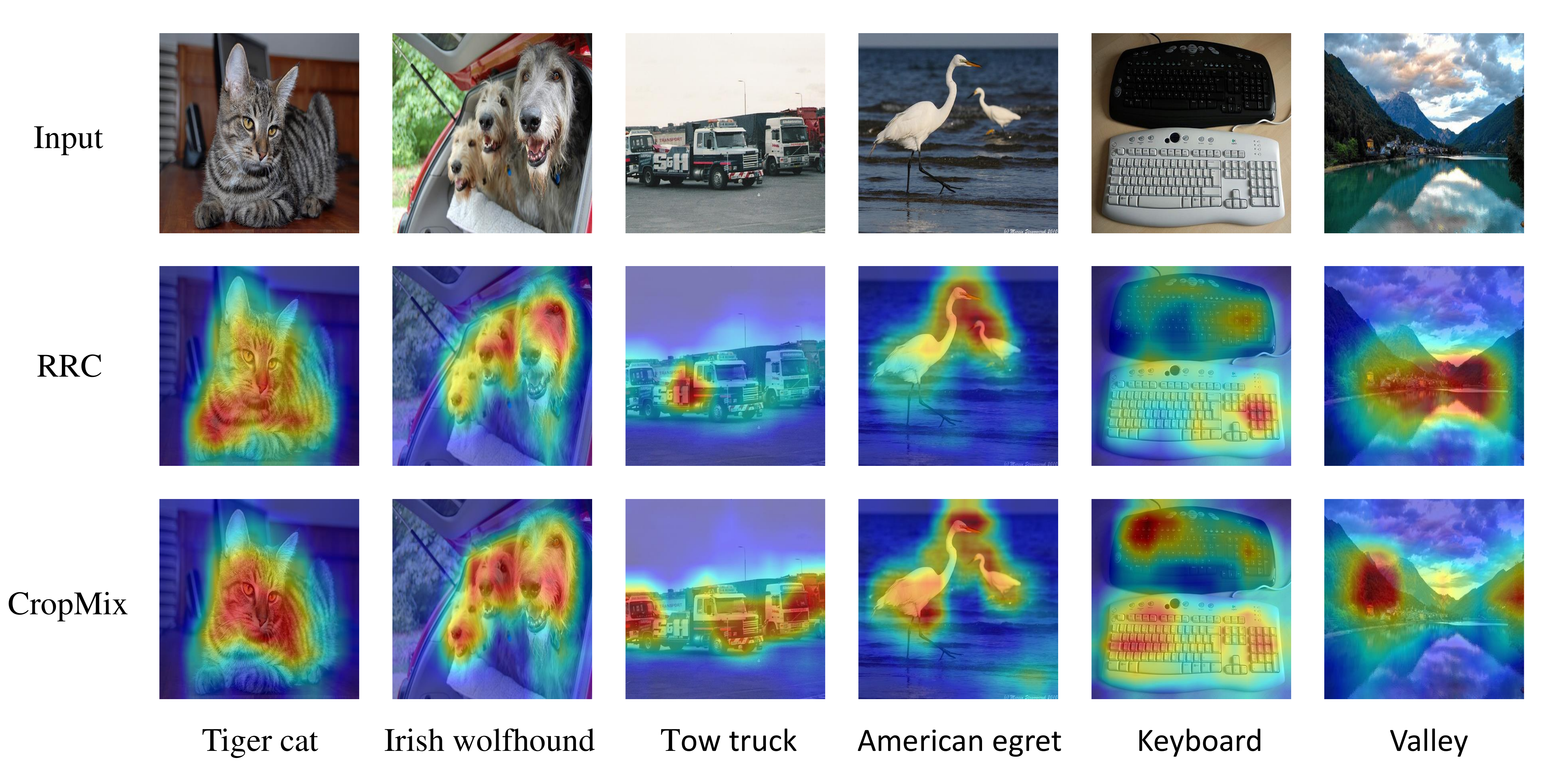}
     \caption{\textbf{Grad-CAM visualization}. We present the Grad-CAM results of RRC and CropMix. }
     \label{fig:grad}
\end{figure*}

\textbf{CAM visualization.} We present the results of applying Grad-CAM~\cite{selvaraju2017grad} on ResNet50 trained with CropMix and RRC. The training recipe is R1.  Qualitative results are presented in Figure~\ref{fig:grad}. 

Compared to RRC, CropMix tends to focus more on larger spatially distributed regions instead of local regions. CropMix also attends to localizing multiple objects, as shown in Tow truck, American egret, and Keyboard. Such a property is not always beneficial, as CropMix may fail to focus on structurally comprehensive regions, as presented in Valley. 

\begin{figure*}[!htb]
     \centering
     \includegraphics[width = 14cm]
     {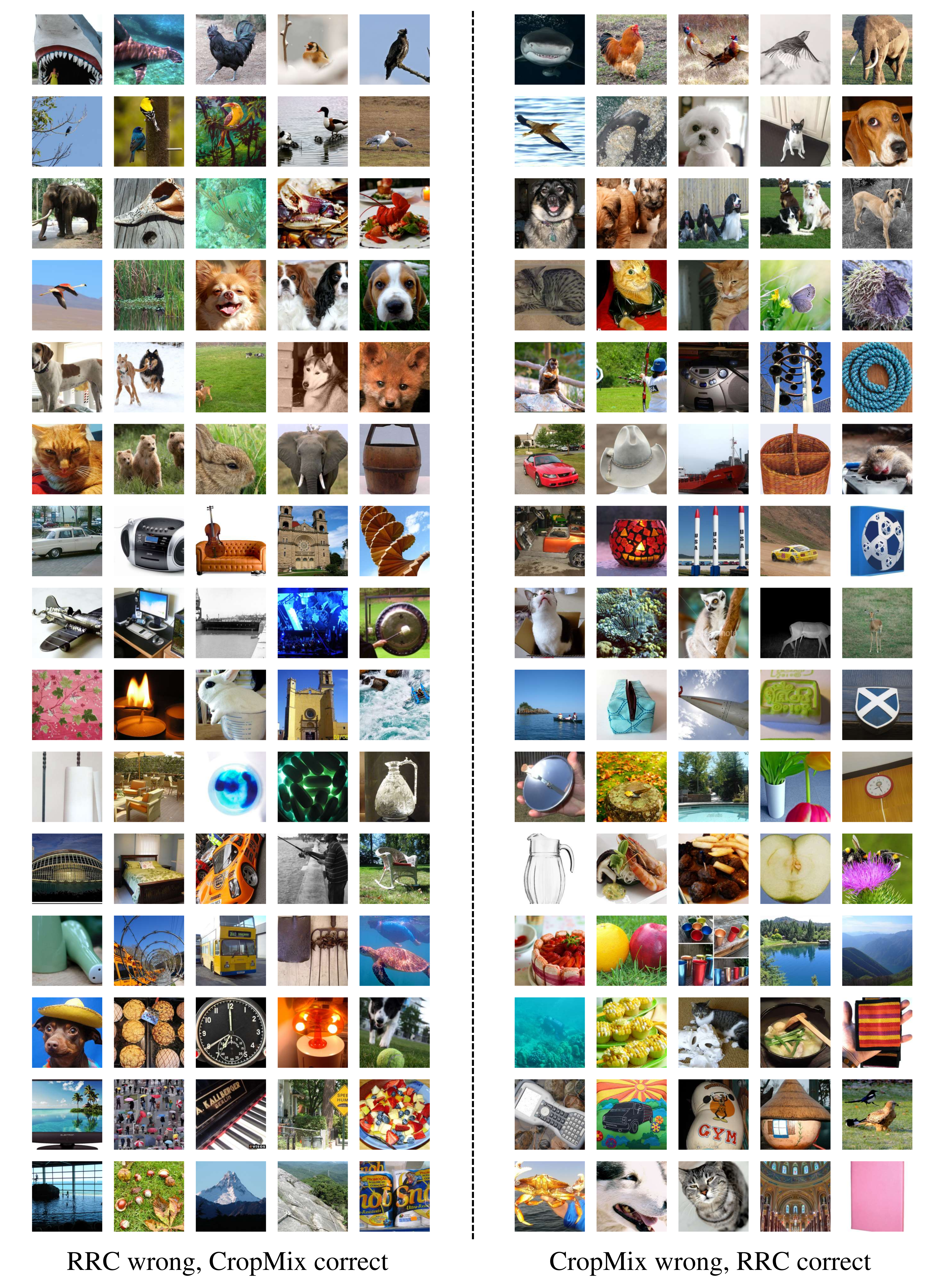}
     \caption{\textbf{Differently classified images.} We show uncurated ImageNet validation images that are differently classified by models trained with RRC and CropMIx.}
     \label{fig:gallery}
\end{figure*}

\textbf{Differently classified images.} 
We present randomly selected images that are differently classified by classifiers trained with RRC and CropMix. The training recipe is R1 and model is ResNet-50. 
Presented images are randomly taken from ImageNet validation set and images with private information are discarded. Visualizations are presented in Figure~\ref{fig:gallery}.

\section{Border impacts}
CropMix improves the performance of image classifiers, either trained from scratch or fine-tuned from pre-trained weights. Improved performance may result in more robust and reliable machine learning systems for real-world applications. However, as no theoretical foundation is developed, the interpretability of CropMix is limited and hence needs to be carefully applied to sensitive applications, such as medical image recognition and self-driving.   Additionally, the simplicity of CropMix may result in lower costs to machine learning practitioners.

\end{document}